\documentclass[11pt,a4paper]{article}
\usepackage[hyperref]{acl2017}
\usepackage{times}
\usepackage{latexsym}

\usepackage{caption}
\usepackage{algorithm}
\usepackage{algorithmicx}
\usepackage{algpseudocode}

\usepackage{multirow}
\usepackage{tikz}
\usepackage{pgfplots}
\usepackage{array}
\usepackage{amsmath}
\usepackage{color}

\usepackage{graphicx}
\usepackage{caption}
\usepackage{subcaption}

\usepackage{url}

\aclfinalcopy 

\title{Encoder-Decoder Shift-Reduce Syntactic Parsing}

\author{Jiangming Liu \and Yue Zhang\\
	    Singapore University of Technology and Design,\\
	    8 Somapah Road, Singapore, 487372\\
	    {\tt \{jiangming\_liu, yue\_zhang\}@sutd.edu.sg}}
	    
\date{}

\begin{document}

\maketitle

\begin{abstract}
Starting from NMT, encoder-decoder neural networks have been used for many NLP problems.
Graph-based models and transition-based models borrowing the encoder components achieve state-of-the-art performance on dependency parsing and constituent parsing, respectively.
However, there has not been work empirically studying the encoder-decoder neural networks for transition-based parsing.
We apply a simple encoder-decoder to this end, achieving comparable results to the parser of \newcite{dyer:2015} on standard dependency parsing, and outperforming the parser of \newcite{vinyals:2015} on constituent parsing.
\end{abstract}

\section{Introduction}
Neural networks have achieved the state-of-the-art for parsing under various grammar formalisms, including dependency \cite{dozat2016deep}, constituent \cite{dyer2016recurrent} and CCG parsing \cite{xu2016expected}.
\newcite{henderson2004discriminative} are the first to apply neural network to parsing. 
The work of \newcite{chen:2014} is seminal to employs transition-based methods in neural network.
This method has been extended by investigating more complex representations of configurations \cite{dyer:2015,ballesteros2015improved} and global training with beam search \cite{zhou2015neural,andor:2016}.
Borrowing ideas from NMT \cite{bahdanau:2015}, recent advances of neural parsing improved performances of both transition-based \cite{kiperwasser2016simple,dyer2016recurrent} and graph-based parsers \cite{kiperwasser2016simple,dozat2016deep}, utilizing a bidirectional RNN as an \textbf{encoder} to represent input sentences.
In particular, using such encoder structure, the graph-based parser of \protect\newcite{dozat2016deep} achieve the state-of-the-art results for dependency parsing.

The success of the encoder structure can be attributed to the use of multilayer bidirectional LSTMs to induce non-local representations of sentences.
Without manual feature engineering, such architecture automatically extracts complex features for syntactic representation. 
For neural machine translation, such encoder structure has been connected to a corresponding LSTM \textbf{decoder}, giving the state-of-the-art for sequence-to-sequence learning.
Compared to the carefully designed stack representations of \newcite{dyer:2015,dyer2016recurrent}, the encoder-encoder structure is simpler, and more general, which can be used across different grammar formalisms without redesigning the stack representation.
\protect\newcite{vinyals:2015} applied the same encoder-decoder structure to constituent parsing, generating the bracketed syntactic trees as the output token sequence.
However, their model achieves relatively low accuracies. 

The advantage of using the decoder is that it leverages the LSTM structure to capture full sequence information in the output.
Unlike greedy or CRF decoders \cite{durrett2015neural}, which capture only local label dependencies, LSTM decoder models global label sequence relations.
One possible reason for the low accuracies of \protect\newcite{vinyals:2015} can be the output sequence, which is a simple bracketed representation of constituent trees, without carefully designed representation of structural correlations between each token.
As a result, strong constraints are necessary to ensure that the output string corresponds to a valid tree \cite{vinyals:2015}.
In contrast, transition-based systems use sequences of shift-reduce actions to build the parse tree, where the actions have intrinsic structural relations. 

Motivated by the above, we study the effectiveness of a very simple encode-decoder structures for shift-reduce parsing.
Our model can be regarded as direct application of the standard neural machine translation architecture to shift-reduce parsing, which is invariant to different grammar formalisms.
In particular, the encoder is used to represent the input sentence and the decoder is used to generate a sequence of transition actions for constructing the syntactic structure.
We additionally use the attention mechanism over the input sequence \cite{vinyals:2015}, but with a slight modification, taking separate attentions to represent the stack and queue, respectively.
On standard PTB evaluation, Our final model achieves 93.1\% UAS for dependency parsing, which is comparable to the model of \protect\newcite{dyer:2015}, and 90.5 on constituent parsing, which is 2.2\% higher compared to \protect\newcite{vinyals:2015}.
We release our source code at \url{https://github.com/LeonCrashCode/Encoder-Decoder-Parser}.

\section{Transition-based parsing}
Transition-based parsers scan an input sentence from left to right, incrementally performing a sequence of transition actions to predict its parse tree.
Partially-constructed outputs are maintained using a stack, and the incoming words are ordered in a queue.
The initial state consists of an empty stack and a queue containing the whole input sentence.
At each step, a transition action is taken to consume the input and construct the output.
The process repeats until the input queue is empty and the stack contains only one element, e.g. a \textit{ROOT} for dependency parsing, and $S$ for constituent parsing and CCG parsing.

\begin{figure}
\begin{center}
\includegraphics[width=8 cm,height=2.7cm]{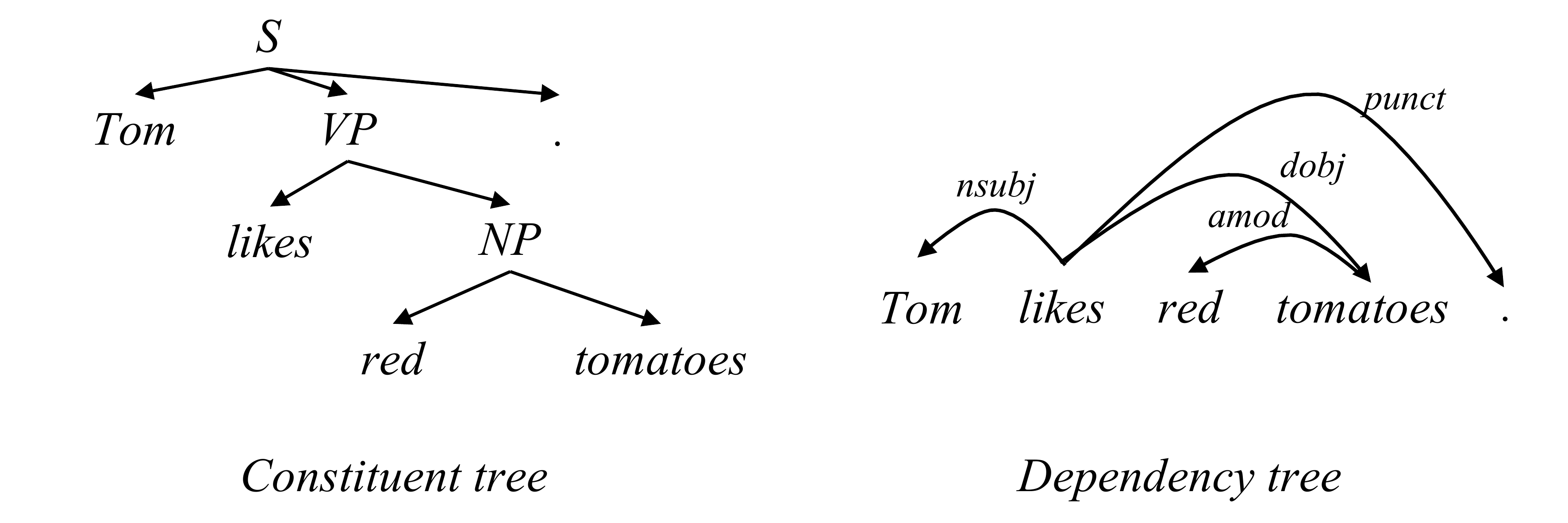}
\end{center}
\caption{\label{examples} Constituent structure and dependency structure of the sentence ``Tom likes red tomatoes."}
\end{figure}

In this paper, we investigate dependency parsing and constituent parsing, which are shown in Figure \ref{examples}, respectively.
As can be seen in the figure, the two formalisms render syntactic structures from very different perspectives.
Correspondingly, the stack structures for transition-based dependency parsing and constituent parsing are very different.
For dependency parsing, the stack contains words directly, while for constituent parsing, the stack contains constituent nodes, which cover spans of words in a sentence. 
In addition, the set of transition actions for building dependency and constituent structures are highly different, as shown by the examples in sections 2.1 and 2.2, respectively.
Traditional approaches, such as the stack LSTM of \newcite{dyer:2015,dyer2016recurrent}, build different representation of the stack for dependency and constituent parsing.
In contrast, our method is agnostic to the stack structure, using an encoder-decoder structure to ``translation" input sentences to output sequences of shift-reduce actions.
To this term, each grammar formalism is reminiscent of a unique foreign language. 

\subsection{Dependency parsing}
We employ the arc-standard transition system \cite{nivre2007maltparser}.
Formally, a parsing state is denoted as $(S,Q,L)$, where $S$ is the stack $[..., s_2, s_1, s_0]$, $Q$ is the queue containing coming words, and $L$ is a set of dependency arcs that have been built.
At each step, the parser chooses one of the following actions:
\begin{itemize}
\item \textsc{Shift}: pop the front word off the queue, and push it onto the stack. 
\item \textsc{Left-Arc($l$)}: add an arc with label $l$ between the top two trees on the stack $(s_1 \leftarrow s_0)$ and remove $s_1$ from the stack.
\item \textsc{Right-Arc($l$)}: add an arc with label $l$ between the top two trees on the stack $(s_1 \rightarrow s_0)$ and remove $s_0$ from the stack.
\end{itemize} 
The arc-standard parser can be summarized as the deductive system in Figure \ref{dependency}.
For a sentence with size $n$, parsing stops after performing exactly $2n-1$ actions. 
Given a sentence of Figure \ref{examples}, the sequence of actions \textsc{Shift}, \textsc{Shift},  \textsc{Left-Arc($nsubj$)},  \textsc{Shift},  \textsc{Shift}, \textsc{Left-Arc($amod$)}, \textsc{Right-Arc($dobj$)}, \textsc{Shift},  \textsc{Right-Arc($punct$)}, can be used to construct its dependency tree. 

\begin{figure}[!tp]
\begin{subfigure}[b]{0.4\textwidth}
\centering
\renewcommand{\arraystretch}{0.8}
\begin{tabular}{>{\small}c>{\small}c}
Initial State & $(\phi, Q, \phi)$\\
Final State & $(s_0, \phi, L)$\\
\\
\multicolumn{2}{>{\small}c}{Induction Rules:} \\
\textsc{Shift} & {\Large$\frac{(S, q_0|Q, L)}{(S|q_0,Q,L)}$} \\
\\
\textsc{Left-Arc(l)} & {\Large$\frac{(S|s_1|s_0, Q, L)}{(S|s_0,Q, L\cup{s_1 \leftarrow s_0})}$} \\
\\
\textsc{Right-Arc(l)} & {\Large$\frac{(S|s_1|s_0, Q, L)}{(S|s_1,Q, L\cup{s_1 \rightarrow s_0})}$} \\
\end{tabular}
\caption{Arc-standard dependency parsing.}
\label{dependency} 
\end{subfigure}

\;
\;
\;
\;
\;
\;
\;

\begin{subfigure}[b]{0.4\textwidth}
\centering
\renewcommand{\arraystretch}{0.8}
\begin{tabular}{>{\small}c>{\small}c}
Initial State & $(\phi, Q, 0)$\\
Final State & $(s_0, \phi, 0)$\\
\\
\multicolumn{2}{>{\small}c}{Induction Rules:} \\
\textsc{Shift} & {\Large$\frac{(S, q_0|Q, n)}{(S|q_0,Q,n)}$} \\
\\
\textsc{NT(X)} & {\Large$\frac{(S, Q, n)}{(S|e(x), Q, n+1)}$} \\
\\
\textsc{Reduce} & {\Large$\frac{(S|e(x)|s_j|...|s_0, Q, n)}{(S|e(x,s_j,...,s_0), Q, n-1)}$} \\
\end{tabular}
\caption{Top-down constituent parsing.}
\label{constituency} 
\end{subfigure}
\caption{Deduction systems}
\label{Deduction}
\end{figure}

\subsection{Constituent parsing}
We employ the top-down transition system of \newcite{dyer2016recurrent}.
Formally, a parsing state is denoted as $(S,Q,n)$, where $S$ is the stack $[..., s_2, s_1, s_0]$ where each element could be a open nonterminal\footnote{An open nonterminal in top-down parsing is an nonterminal waiting to be completed}, a completed constituent, or a terminal, $Q$ is the queue, and $n$ is the number of open nonterminals on the stack.
At each step, the parser chooses one of the following actions:
\begin{itemize}
\item \textsc{Shift}: pop the front word off the queue, and push it onto the stack. 
\item \textsc{NT(X)}: open a nonterminal with label X on top of the stack. 
\item \textsc{Reduce}: repeatedly pop completed subtrees or terminal symbols from the stack until an open nonterminal is encountered, and then this open NT is popped and used as the label of a new constituent that has the popped subtrees as its children. This new completed constituent is pushed onto the stack as a single composite item.
\end{itemize}
The top-down parser can be summarized as the deductive system in Figure \ref{constituency}.
Given the sentence in Figure \ref{examples}, the sequence of actions \textsc{NT($S$)}, \textsc{Shift}, \textsc{NT($VP$)},  \textsc{Shift},  \textsc{NT($NP$)}, \textsc{Shift}, \textsc{Shift}, \textsc{Reduce}, \textsc{Reduce}, \textsc{Shift}, \textsc{Reduce}, can be used to construct its constituent tree. 

\subsection{Generalization}
Both transition systems above can be treated as examples of a general sequence-to-sequence task.
Formally, given a sentence $x_1, x_2, ..., x_n$ where $x_i$ is the $i$th word in the sentence, the goal is to generate a corresponding sequence of actions $a_1, a_2, ..., a_m$, which correspond to a syntactic structure.
Other shift-reduce parser systems, such as CCG, can be regarded as instances of this. 

\begin{figure}
\begin{center}
\includegraphics[width=8cm,height=3.7cm]{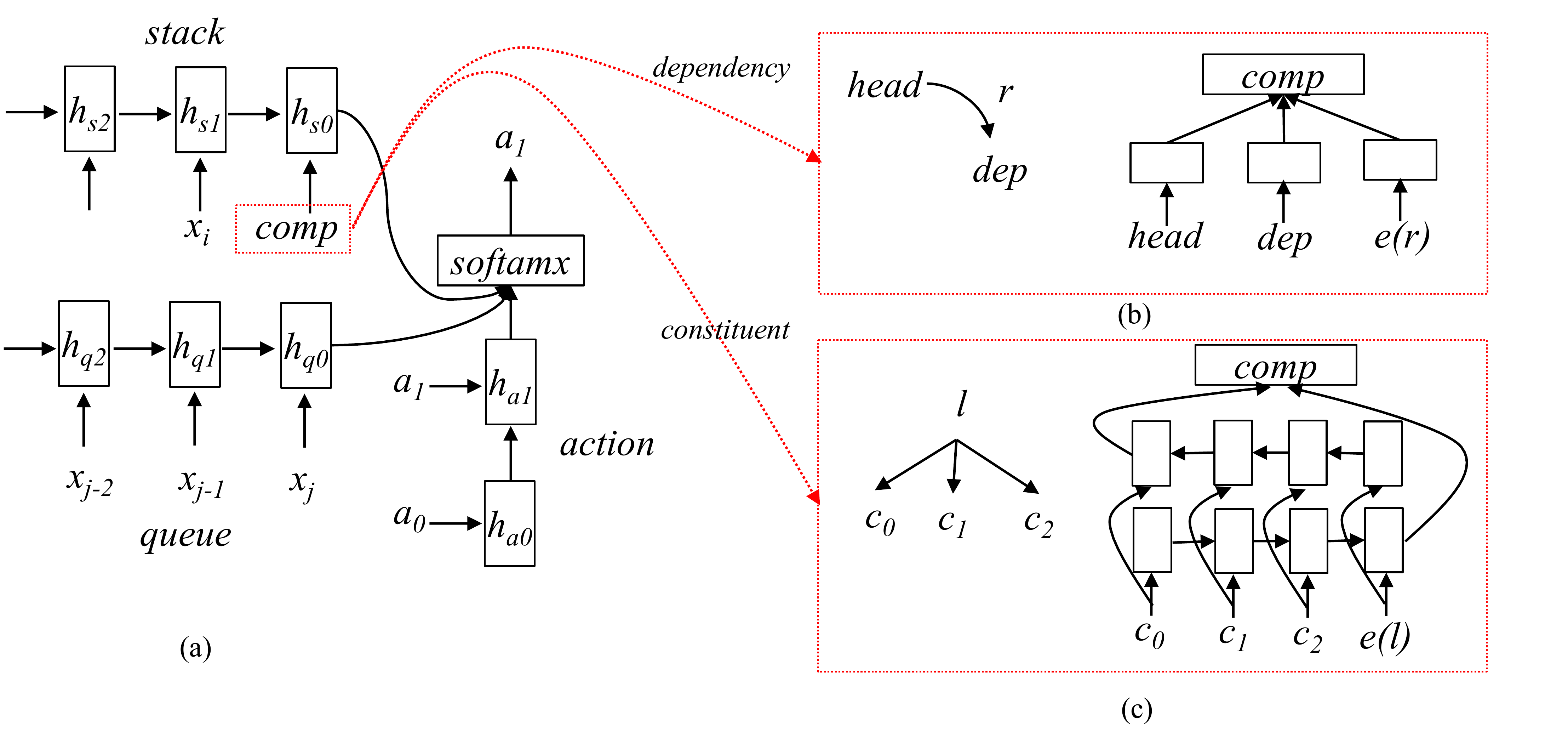}
\end{center}
\caption{\label{stack-lstm} Structure of stack-LSTM with dependency and constituent composition, respectively.}
\end{figure}

\section{Baseline}
We take two baseline neural parsers, namely the parser of \newcite{dyer:2015,dyer2016recurrent} and the parser of \newcite{vinyals:2015}.
The former is used to study the effect of our formalism-independent representation, while the latter can be used to demonstrate the advantage of transition-based system and the encoder-decoder framework.
We briefly describe the parsers of \protect\newcite{dyer:2015,dyer2016recurrent} in this section, and the structure of \newcite{vinyals:2015} in Sections 4.1 and 4.2.

As shown in Figure \ref{stack-lstm}(a), the parser of \newcite{dyer:2015} has three parts: 1) a stack of partial output, implemented using a stack-LSTM, 2) a queue of incoming words using an LSTM and 3) a list of actions that has been taken so far encoded by an LSTM.
The stack-LSTM is implemented left to right, the queue LSTM is implemented right to left, and the action history LSTM in the first-to-last order.
The last hidden states of each LSTM is concatenated and fed to a softmax layer to determine the next action given the current state:
\begin{equation*}
p(act) = \textit{softmax}(W [h_s;h_q;h_a]+b),
\end{equation*}
where $h_s$, $h_q$ and $h_a$ denotes the last hidden states of the stack LSTM, the queue LSTM and the action history LSTM, respectively.

The stack-LSTM parser represents states on the stack by task-specific composition functions.
We give the composition functions for dependency parsing \cite{dyer:2015} and constituent parsing \cite{dyer2016recurrent}, respectively below.

\textbf{Dependency parsing} ~~~ 
The composition function models the dependency arc between a head and its dependent, i.e. $head \overset{r}{\rightarrow} dep$, when a \textsc{Reduce} action is applied, as shown in Figure \ref{stack-lstm}(b):
\begin{equation*}
comp = tanh(W_{comp} [h_{s_{head}};h_{s_{dep}}; e(r)]+b_{comp}),
\end{equation*}
where $h_{s_{h}}$ is the value of the head, $h_{s_{d}}$ is the value of the dependent and $e(r)$ is the arc relation embedding. 
After a \textsc{Left-Arc($r$)} action is taken, $h_{s_{h}}$ and $h_{s_{d}}$ are removed from the stack-LSTM, and then $comp$ is push onto the stack-LSTM.

\textbf{Constituent parsing} ~~~ 
The composition function models the constituent spanning their children, i.e. ($l$ ($c2$) ($c1$) ($c0$)), when a \textsc{Reduce} action is applied, as shown in Figure \ref{stack-lstm}(c):
\begin{equation*}
comp = \textsc{Bi-LSTM}_{comp}([h_{s_{c2}},h_{s_{c1}}, h_{s_{c0}}, e(l)]),
\end{equation*}
where $h_{s_{c2}}$, $h_{s_{c1}}$ and $h_{s_{c0}}$ are the value of the children on stack, and $e(l)$ is the constituent label embedding. 
After a \textsc{Reduce} action is taken, $h_{s_{c2}}$, $h_{s_{c1}}$ and $h_{s_{c0}}$ are removed from the stack-LSTM, and then $comp$ is push onto the stack-LSTM.

It is worth noting that the stack contains similar information compared to action history.
This is because the content of the stack can be inferred when the action history is given.
As a result, the stack structure of the parser by \newcite{dyer:2015} is redundant; it only serves as a different way of extracting features given a sequence of actions that have been applied.
Our parser models only the action sequence, relying on the model to infer necessary information about the stack automatically.
\begin{figure*}
\begin{center}
\includegraphics[width=16.01cm,height=4.01cm]{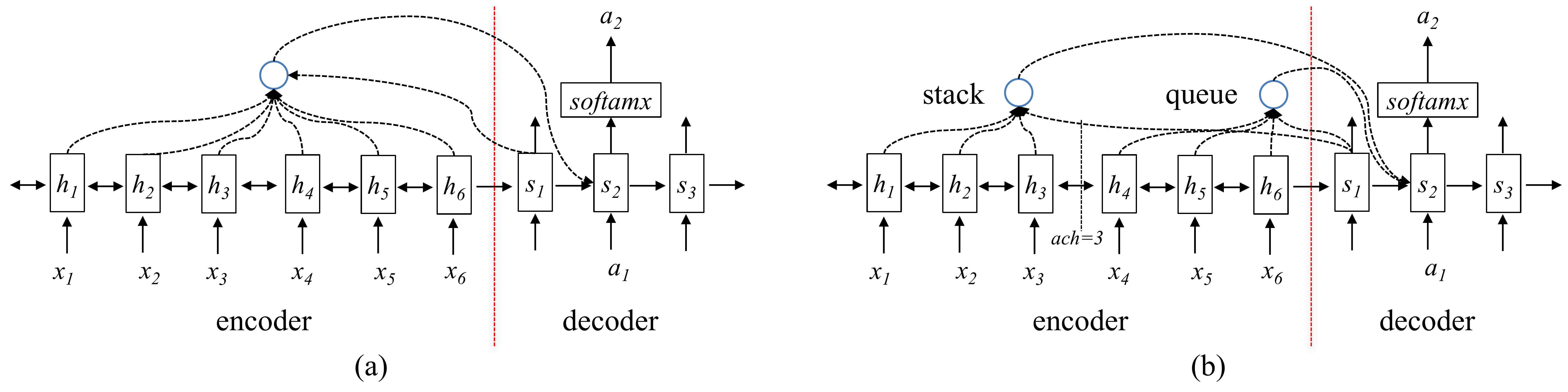}
\end{center}
\caption{\label{structure} Encoder-decoder structure for parsing. (a) vanilla decoder; (b) Stack-queue decoder, where the stack and the queue are differentiated by $ach$, which is initialized to the beginning of the sentence ($ach = 0$), meaning the stack is empty and queue contains the whole sentence.}
\end{figure*}

\section{Model}
As shown in Figure \ref{structure}, our model structure consist of two main components, namely encoder and decoder.
The encoder is a bidirectional recurrent neural network, representing information of the input sentence; the decoder is a different recurrent neural network, used to output a sequence of transition actions. 
The encoder can be further divided into two parts, which contain words of stack and queue, respectively for transition-based parsing.

\subsection{Encoder}
We follow \newcite{dyer:2015}, representing each word using three different types of embeddings including pretrained word embedding, $\overline{e}_{w_i}$, which is not fine-tuned during training of the parser, randomly initialized embeddings $e_{w_i}$, which is fine-tuned, and the randomly initialized part-of-speech embeddings, which is fine-tuned.
The three embeddings are concatenated, and then fed to nonlinear layer to derive the final word embedding: 
\begin{equation*}
x_i = f(W_{enc} [e_{p_i}; \overline{e}_{w_i}; e_{w_i}]+b_{enc}),
\end{equation*}
where $W_{enc}$ and $b_{enc}$ are model parameters, $w_i$ and $p_i$ denote the form of the pos $i$th input word, respectively, and $f$ is an nonlinear function.
In this paper, we use ReLu for $f$. 

The encoder is based on bidirectional peephole connected LSTM \cite{greff2016lstm}, which takes sequence of the word embeddings $x_i$ as input, and output the sequence of hidden state $h_i$.
Bi-LSTM is adopted in our models:
\begin{equation*}
h_i = [h_{l_i};h_{r_i}] = \textsc{Bi-LSTM}(x_i).
\end{equation*}
The sequence of $h_i$ is fed to the decoder.

\subsection{Vanilla decoder}
As shown in Figure \ref{structure}(a), the decoder structure is similar to that of neural machine translation.
It applies an LSTM to generate sequences of actions:
\begin{equation*}
s_j= g(W_{dec}[s_{j-1};e_{a_{j-1}};h_{att_{j}}]+b_{dec}),
\end{equation*}
where $W_{dec}$ and $b_{dec}$ are model parameters, $a_{j-1}$ is previous action, $e_{a_{j-1}}$ is the embedding of $a_{j-1}$, $s_{j-1}$ is the LSTM hidden state for $a_{j-1}$, and $s_j$ is the current LSTM state, from which $a_j$ is predicted.
$h_{att_j}$ is the result of attention over the encoder states $h_1...h_{n}$ using the $j$th decoder state:
\begin{equation*}
h_{att_j}=\textsc{attention}(1,n)= \sum_{i=1}^{n}{\alpha_ih_i}
\end{equation*}
where
\begin{equation*}
\alpha_i = \frac{exp(\beta_i)}{\sum^n_{k=1}{exp(\beta_k)}},
\end{equation*}
and the weight scores $\beta$ are calculated by using the previous hidden state $s_{j-1}$ and corresponding encoder hidden state $h$:
\begin{equation*}
\beta_i = U^Ttanh(W_{att}\cdot[h_i; s_{j-1}] + b_{att}).
\end{equation*}
$s_j$ is used to predict current action $a_j$:
\begin{equation*}
p(a_j|s_j) = \textit{softmax}(W_{out} * s_j + b_{out})).
\end{equation*}
Here $W_{att}$, $b_{att}$, $W_{out}$, $b_{out}$ are model parameters, $g$ is nonlinear function, we use the ReLu for $g$.
For the encoder, the initial hidden state are randomly initialized model parameters; For the decoder, the initial LSTM state $s_0$ is the last the encoder hidden state $[h_{l_n};h_{r_1}]$.

This vanilla encoder decoder structure is identical to the method of \protect\newcite{vinyals:2015}.
The only difference is that we use shift-reduce action as the output, while \protect\newcite{vinyals:2015} use bracketed string of constituent trees as the output.

\subsection{Stack-Queue decoder}
We extend the vanilla decoder, using two separate attention models over encoder hidden state to represent the stack and queue, respectively, as shown in Figure \ref{structure}(b).
In particular, for a given state, the encoder is divided into two segments, with the left segment (i.e. \textit{stack segment}) containing words form $x_1$ to the word on top of the stack $x_t$, and the right segment (i.e. \textit{queue segment}) containing words from the front of the queue $x_{t+1}$ to $x_n$



\textbf{Attention} is applied to the stack and the queue segments, respectively.
In particular, the representation of the stack segment is:
\begin{equation*}
h_{l_{att_j}}=attention(1,t)= \sum_{i=1}^{t}{\alpha_ih_i},
\end{equation*}
and the representation of the queue segment is:
\begin{equation*}
h_{r_{att_j}}=attention(t+1,n)= \sum_{i=t+1}^{n}{\alpha_ih_i}.
\end{equation*}
Similar with the vanilla decoder, the hidden unit is:
\begin{equation*}
s_j= g(W_{dec}[s_{j-1};e_{a_{j-1}};h_{l_{att_j}};h_{r_{att_j}}]+b_{dec}).
\end{equation*}
Where $g$ is the same nonlinear function as in vanilla decoder.

\subsection{Training}
Our models are trained to minimize a cross-entropy loss objective with an $l_2$ regularization term, defined by
\begin{equation*}
L(\theta)=-\sum_i{\sum_j{log~p_{a_{ij}}}} + \frac{\lambda}{2}||\theta||^2,
\end{equation*}
where $\theta$ is the set of parameters, $p_{a_{ij}}$ is the probability of the $j$th action in the $i$th training example given by the model and $\lambda$ is a regularization hyper-parameter. $\lambda = 10^{-6}$.
We use stochastic gradient descent with Adam to adjust the learning rate.

\section{Experiments}
\subsection{Data}
We use the standard benchmark of WSJ sections in PTB \cite{Marcus:1993wd}, where the sections 2-21 are taken for training data, section 22 for development data and section 23 for test for both dependency parsing and constituent parsing.
For dependency parsing, the constituent trees in PTB are converted to Stanford dependencies (version 3.3.0) using the Stanford parser\footnote{https://nlp.stanford.edu/software/lex-parser.shtml}.
We adopt the pretrained word embeddings generated on the AFP portion of English Gigaword \cite{dyer:2015}.

\subsection{Hyper-parameters}
The hyper-parameter values are chosen according to the performance of the model on the development data for dependency parsing, and final values are shown in Table \ref{parameter}.
For constituent parsing, we use the same hyper-parameters without further optimization.

\begin{table}[!tp]
\begin{center}
\renewcommand{\arraystretch}{0.8}
\begin{tabular}{>{\small}l|>{\small}c}
\bf Parameter & Value \\
\hline
\hline
Encoder LSTM Layer & 2 \\
Decoder LSTM Layer & 1 \\
Word embedding dim & 64 \\
Fixed word embedding dim & 100 \\
POS tag embedding dim & 6 \\
Label embedding dim & 20 \\
Action embedding dim & 40 \\
encoder LSTM input dim & 100 \\
encoder LSTM hidden dim & 200 \\
decoder LSTM hidden dim & 400 \\
Attention hidden dim & 50 \\
\hline
\end{tabular}
\end{center}
\caption{\label{parameter} Hyper-parameters.}
\end{table}

\subsection{Development experiments}
Table \ref{dev_dep} shows the development results on dependency parsing.
To verify the effectiveness of attention, we build a baseline using average pooling instead (SQ decoder + average pooling).
We additionally build a baseline (SQ decoder + treeLSTM) that is aware of stack structures, by using a tree-LSTM \cite{tai2015improved} to derive head node representations when dependency arcs are built.
Attention on the stack sector are applied only on words on the stack, but not for their dependents.
This representation is analogous to the stack representation of \protect\newcite{dyer:2015} and \protect\newcite{watanabe2015transition}.

Results show that the explicit construction of stack does not bring significant improvements over our stack-agnostic attention model, which confirms our observation in Section 3 that the action history information is sufficient for inferring the stack structure.
Our model achieved this goal to some extent. 
The SQ decoder with average pooling achieves a 3.4\% UAS improvement, compared to the vanilla decoder (Section 4.2).
The SQ decoder with attention achieves a further 0.5\% UAS improvement, reaching comparable results to the stack-LSTM parser.

\begin{table}[!tp]
\begin{center}
\renewcommand{\arraystretch}{0.8}
\begin{tabular}{>{\small}l|>{\small}c}
\bf Model & UAS (\%) \\
\hline
\hline
\newcite{dyer:2015} & 92.3\\
\hline
Vanilla decoder & 88.5 \\
SQ decoder + average pooling& 91.9 \\
SQ decoder + attention & 92.4 \\
SQ decoder + treeLSTM & 92.4 \\
\end{tabular}
\end{center}
\caption{\label{dev_dep} The development results for dependency parsing.}
\end{table}

\subsection{Comparison to stack-LSTM}
We take a range of different perspectives to analysis the errors distribution of our parser, comparing them with stack-LSTM parser \cite{dyer:2015}.
The parsers show different empirical performances over these measures.

Figure \ref{dep_len} shows the accuracy of the parsers relative to the sentence length.
The parsers perform comparatively better in short sentences.
The stack-LSTM parser performs better on relatively short sentences ($\leq$ 30), while our parser performs better on longer sentences.
The composition function is applied in the stack-LSTM parser to explicitly represent the partially-constructed trees, ensuring high precision of short sentences.
On the other hand, errors are also fully represented and accumulated in long sentences.
As the sentence grows longer, it is difficult to capture the stack structure.
With stack-queue sensitive attention, SQ decoder implicitly represent the structures.
The decoder is used to model sequences of actions globally, and is less influenced by error propagation.

\begin{figure}[!tp]
\begin{tikzpicture}[scale=1, font=\small]
\begin{axis} [
height = 5cm,
width = 8cm,
xlabel = Sentence length,
ylabel = Dependency accuracy (\%),
xmin=10,
xmax=60,
ymin=87,
ymax=95,
ytick pos=left,
legend style={at={(0.5,0.5)}},
ymajorgrids=true,
grid style=dashed,
y label style={at={(0.05,0.5)}}
]
\addplot[thick] coordinates{
	(10, 93.8)
	(20, 94.5)
	(30, 93.4)
	(40, 92.8)
	(50, 91.6)
	(60, 87.6)
};
\addlegendentry{stack-LSTM}
\addplot[dashed] coordinates{
	(10, 93.6)
	(20, 94.3)
	(30, 93.3)
	(40, 93.1)
	(50, 91.8)
	(60, 89.0)
};
\addlegendentry{SQ decoder}
\end{axis}
\end{tikzpicture}
\caption{Accuracy against sentence length. (the number of words in a sentence, in bins of size 10, where 20 contains sentences with length [10, 20).)}
\label{dep_len} 
\end{figure}
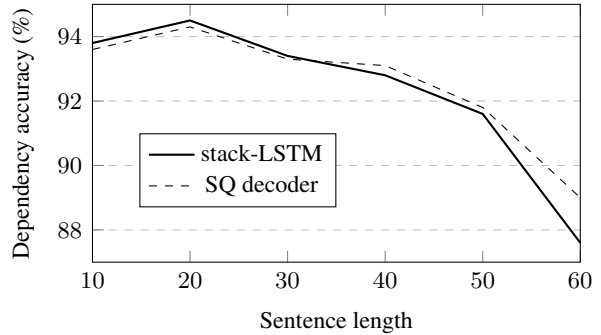

\begin{figure*}[!tp]
\begin{center}
\begin{tikzpicture}[scale=1, font=\small]
\begin{axis}[
height = 4.0cm,
width = 16cm,
    ybar,
    ymin= 86,
    ymax = 98,
    bar width=5pt,
    legend style={at={(0.68,1)}, anchor=north},
    ylabel = {constituents recall (\%)},
    y label style={at={(0.03,0.4)}},
    x label style={at={(0.5,0.0)}},
    grid style=dashed,
    ymajorgrids=true,
    symbolic x coords={
    NN,
    IN,
    NNP,
    DT,
    JJ,
    NNS,
    RB,
    CD,
    VBD,
    VB,
    CC,
    TO,
    VBZ,
    VBN,
    PRP
    },
    xtick=data,
    xticklabel style={
        inner sep=0pt,
        rotate=20
    },
    ]
    \addplot[fill=gray!20]coordinates {
        (NN, 94.4582407528 )
(IN, 87.6611418048 )
(NNP, 95.2953296703 )
(DT, 97.8838174274 )
(JJ, 94.5033482143 )
(NNS, 93.947144075 )
(RB, 86.7283950617 )
(CD, 95.8570688762 )
(VBD, 94.676180022 )
(VB, 91.2427559562 )
(CC, 86.0380116959 )
(TO, 92.0967741935 )
(VBZ, 93.6222403925 )
(VBN, 87.7906976744 )
(PRP, 96.8631178707 )
     };
     \addplot[fill=gray!80] coordinates {
     	(NN, 94.5235916874 )
(IN, 88.1131759585 )
(NNP, 95.5013736264 )
(DT, 97.4481327801 )
(JJ, 95.0892857143 )
(NNS, 93.5493037795 )
(RB, 86.7798353909 )
(CD, 95.64992232 )
(VBD, 93.1942919868 )
(VB, 91.9510624598 )
(CC, 86.6959064327 )
(TO, 92.5 )
(VBZ, 93.0498773508 )
(VBN, 89.0365448505 )
(PRP, 97.433460076 )
     };

     \legend{stack-LSTM, SQ-decoder}
\end{axis}
\end{tikzpicture}
\end{center}
\caption{\label{pos} Accuracy against part-of-the-speech tags.}
\end{figure*}
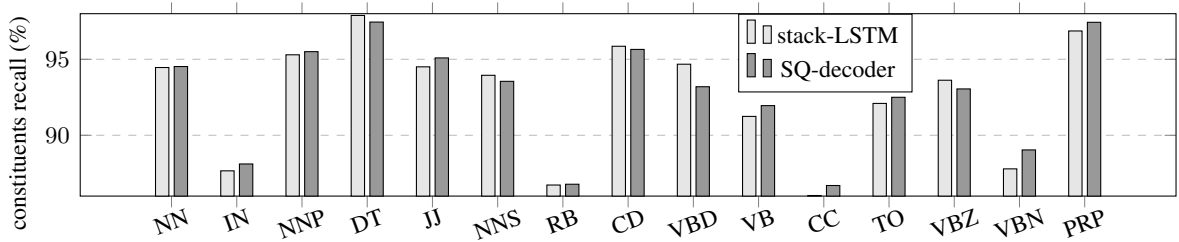

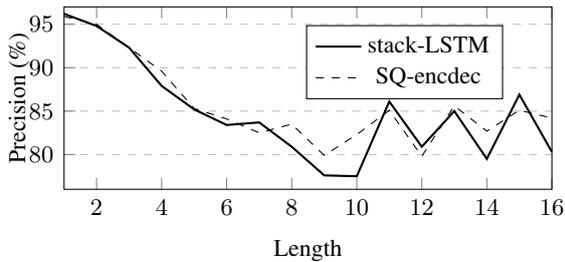
\begin{figure}[!tp]
\begin{center}
\begin{tikzpicture}[scale=1, font=\small]
\begin{axis}[
    	height = 4cm,
	width = 8cm,
	xlabel = {Length},
	ylabel = Precision (\%),
	xmin=1,
	xmax=16,
	ymin=76,
	ymax=97,
	ytick pos=left,
	legend style={at={(0.9,0.9)}},
	ymajorgrids=true,
	grid style=dashed,
	y label style={at={(0.1,0.5)}}
    ]
    \addplot[thick] coordinates{
	(1, 96.2)
	(2, 94.8)
	(3, 92.3)
	(4, 87.9)
	(5, 85.2)
	(6, 83.4)
	(7, 83.7)
	(8, 80.9)
	(9, 77.6)
	(10, 77.5)
	(11, 86.1)
	(12, 80.9)
	(13, 85)
	(14, 79.5)
	(15, 86.9)
	(16, 80.3)
  	};
    \addlegendentry{stack-LSTM}
    \addplot[dashed] coordinates{
	(1, 95.9)
	(2, 95.0)
	(3, 92.3)
	(4, 89.6)
	(5, 85.3)
	(6, 84.1)
	(7, 82.5)
	(8, 83.5)
	(9, 79.9)
	(10, 82.3)
	(11, 85.1)
	(12, 79.8)
	(13, 85.6)
	(14, 82.7)
	(15, 85.1)
	(16, 84.2)
    };

     \legend{stack-LSTM, SQ-encdec}
\end{axis}
\end{tikzpicture}
\end{center}
\caption{\label{arc_len} Arc precision against dependency length. The length is defined as the absolute difference between the indices of the head and modifier.}
\end{figure}

Figures \ref{pos} and \ref{arc_len} show comparison on various POS and dependency lengths, respectively.
While the error distributions of the two parsers on these fine-grained metrics are slightly different, with our model being stronger on arcs that take relatively more steps to build, the main trends of the two models are consistent, which shows that our model can learn similar sources of information compared to the parser of \newcite{dyer:2015}, without explicitly modelling stack information.
This again verifies the usefulness of the decoder on exploiting action history. 

\begin{figure}
\begin{center}
\includegraphics[width=8.01cm,height=5.5cm]{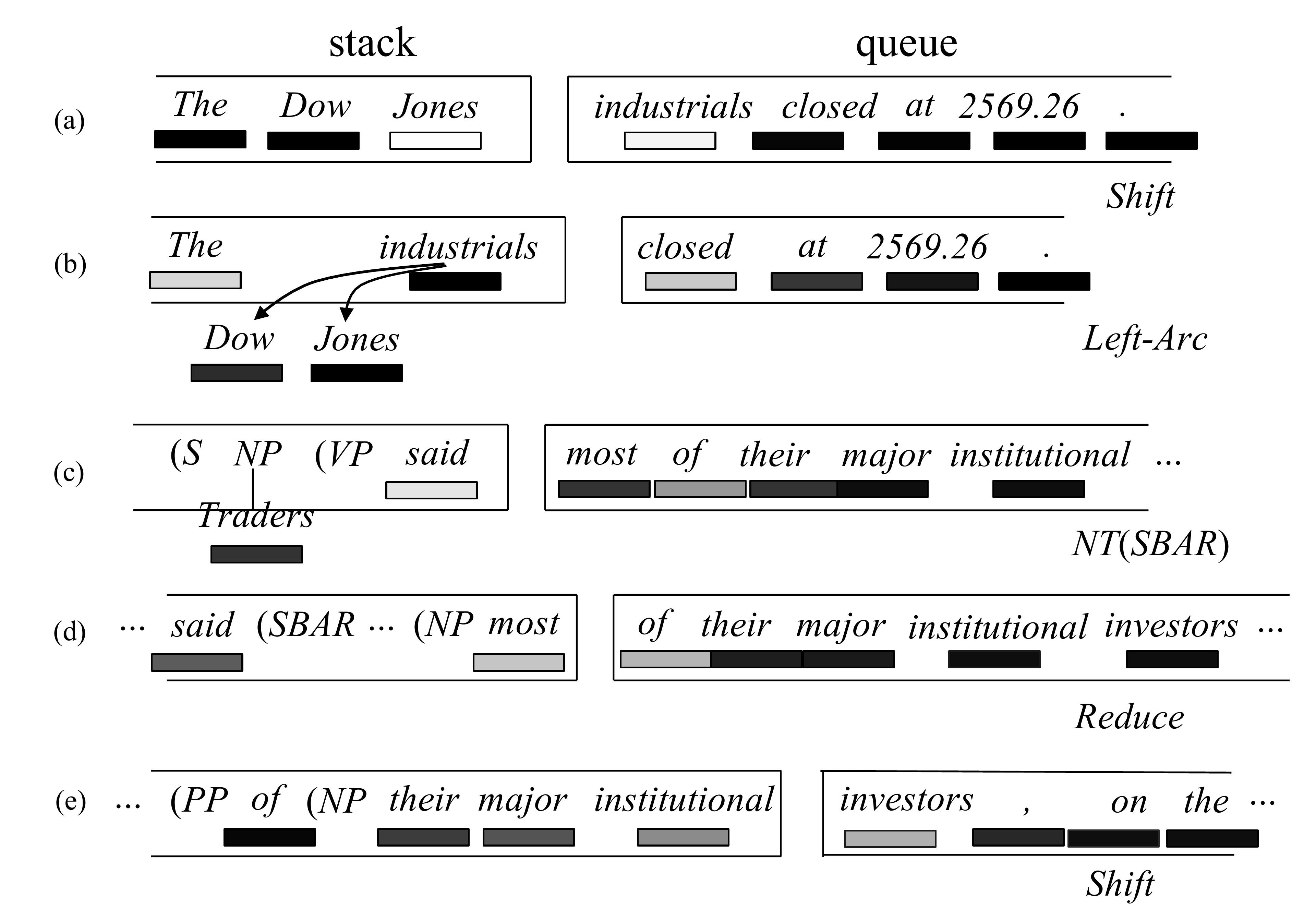}
\end{center}
\caption{\label{example} Output examples to visualize attention values. The grey scale indicates the value of the attention. (a) (b) are for dependency parsing, and (c) (d) (e) are for constituent parsing.}
\end{figure}
\subsection{Attention visualization}
We visualize the attention values during parsing, as shown in Figure \ref{example}.
The parser can implicitly extract the structure features by assigning different attention value to the elements on stack.
In Figure \ref{example}(a), ``Jones" on the top of stack and ``industrials" on the front of queue dominates the prediction of \textsc{Shift} action.
In Figure \ref{example}(b), ``The" on the top of stack and ``closed" on the front of queue contribute more to the prediction of \textsc{Left-Arc}, which constructs an left arc from ``industrials" to ``The" to complete dependency of the word ``industrials". 
In Figure \ref{example}(c), ``said"  on the top of stack determines the prediction of \textsc{NT(\textit{SBAR})} for a clause.
In Figure \ref{example}(d), ``of" on the front of queue suggests to complete the noun phrase of ``most". 
In Figure \ref{example}(e), ``their major institutional" on top of the stack needs the word ``investor" on the front of queue to complete a noun phrase.

Interestingly, these attention values capture information no only from nodes on the stack, but also their dependents, achieving similar effects as the manually defined features of \newcite{chen:2014} and \newcite{kiperwasser2016simple}. In addition, the range of features that our attention mechanism models is far beyond the manual feature templates, since words even on the bottom of the stack can sometimes influence the decision, as shown in Figure \ref{example}(b). These are worth noting given that our model does not explicitly model the stack structure.

\section{Final results}
We compare the final results with previous related work under the fully-supervised setting (except for pretrained word embeddings), as shown in Table \ref{final_dep} for dependency parsing, and Table \ref{final_con} for constituent parsing.
For dependency parsing, our models achieve comparable UAS to the majority of parsers \cite{dyer:2015,kiperwasser2016simple,andor:2016}.

For constituent parsing, our models outperforms the parser of \newcite{vinyals:2015} by differentiating stack and queue and generating transition actions instead.
This shows the advantage of shift-reduce actions over bracketed syntactic trees as decoder outputs.
Using the settings tuned on the dependency development data directly, our model achieves a F1-score of 90.5, which is comparable to the models of \newcite{zhu2013fast} and \newcite{socher2013parsing}.
By using the rerankers of \protect\newcite{choe:2016} under the same settings, we obtain 92.7 F1-score with fully-supervised reranking and 93.4 F1-score with semi-supervised reranking. 

\begin{table}[!tp]
\begin{center}
\renewcommand{\arraystretch}{0.8}
\begin{tabular}{>{\small}l|>{\small}c|>{\small}c}
\bf Model & UAS (\%) & LAS (\%) \\
\hline
\hline
\multicolumn{3}{>{\small}l}{Graph-based}\\
\hline
\newcite{kiperwasser2016simple} & 93.0 & 90.9\\
\newcite{dozat2016deep} & 95.7 & 94.1 \\
\hline
\multicolumn{3}{>{\small}l}{Transition-based}\\
\hline
\newcite{chen:2014} &91.8 & 89.6 \\
\newcite{dyer:2015} & 93.1 & 90.9\\
\newcite{kiperwasser2016simple}$\dagger$ & 93.9 & 91.9\\
\newcite{andor:2016} & 92.9 & 91.0\\
\newcite{andor:2016}* & 94.6 & 92.8\\
\hline
SQ decoder + attention & 93.1 & 90.1\\
\end{tabular}
\end{center}
\caption{\label{final_dep} Results for dependency parsing, where * use global training, $\dagger$ use dynamic oracle.}
\end{table}

\begin{table}[!tp]
\begin{center}
\renewcommand{\arraystretch}{0.8}
\begin{tabular}{>{\small}l|>{\small}c}
\bf Model & F1 (\%) \\
\hline
\hline
\newcite{vinyals:2015} & 88.3 \\
\newcite{socher2013parsing} & 90.4 \\
\newcite{zhu2013fast} & 90.4\\
\newcite{shindo2012bayesian} & 91.1\\
\newcite{dyer2016recurrent} & 91.2\\
\newcite{dyer2016recurrent} -rerank & 93.3\\
\newcite{choe:2016} -rerank& 92.4\\
\hline
SQ decoder + attention & 90.5 \\
SQ decoder + attention -rerank & 92.7 \\
SQ decoder + attention -semi-rerank & 93.4 \\
\end{tabular}
\end{center}
\caption{\label{final_con} Results for constituent parsing.}
\end{table}

\section{Related work}
LSTM encoder structures have been used in both transition-based and graph-based parsing.
Among transition-based parsers, \protect\newcite{kiperwasser2016simple} use two-layer encoder to encode input sentence, extracting 11 different features from a given state in order to predict the next transition action, showing that the encoder structure lead to significant accuracy improvements over the baseline parser of \protect\newcite{chen:2014}.
Among graph-based parsers, \protect\newcite{dozat2016deep} exploit 4-layer LSTM encoder over the input, using conceptually simple biaffine attention mechanism to model dependency arcs over the encoder, resulting in the stat-of-the-art accuracy in dependency parsing.
Their success forms a strong motivation of our work.

The only existing method that directly applies the encoder-decoder structure of NMT to parsing is \protect\newcite{vinyals:2015}, who applied two-lay LSTM for the encoder, and two-layer LSTM decoder to generate bracket syntactic trees.
To our knowledge, we are the first to try a straight forward attention over the encoder-decoder structure for shift-reduce parsing.

\protect\newcite{vinyals:2015} can also be understood as building a language model over bracket constitute trees.
A similar idea is proposed by \protect\newcite{choe:2016}, who directly use LSTMs to model such output forms.
The language model is used to rerank candidate trees from a baseline parser, and trained over large automatically parsing data using tri-training, achieving a current best results for constituent parsing.
Our work is similar in that it can be regarded as a form of language model, over shift-reduce actions rather than bracketed syntactic trees.
Hence, our model can potentially be used for under tri-training settings also. 


There has also been a strand of work applying global optimization to neural network parsing.
\protect\newcite{zhou2015neural} and \protect\newcite{andor:2016} extend the parser of \protect\newcite{zhang2011syntactic}, using beam search and early update training.
They set a max-likelihood training objective, using probability mass in the beam to approximate partition function of CRF training. 
\protect\newcite{watanabe2015transition} study constituent parsing by using a large-margin objective, where the negative example is the expected score of all states in the beam for transition-based parsing.
\protect\newcite{xu2016expected} build CCG parsing models with a training objective of maximizing the expected F1 score of all items in the beam when parsing finishes, under the transition-based system.
More relatedly, \protect\newcite{wiseman2016sequence} use beam search and global max-margin training for the method of \protect\newcite{vinyals:2015}.
In contrast, we use greedy local model; our method is orthogonal to these techniques.

\section{Conclusion and Future work}
We adopted the simple encoder-decoder neural network with slight modification on shift-reduce parsing, achieving comparable results to the current parsers under the same setting.
One advantage of our model is that NMT techniques, such as scheduled sampling \cite{bengio2015scheduled}, residual networks \cite{he2016deep} and ensemble mechanism can be directly applied, which we leave for future work.
Our model can also be trained using tri-training techniques directly, as for \protect\newcite{choe:2016}.
The general encoder-decoder parsing model makes it potentially possible for multi-task training \cite{bahdanau:2015}. 
We will train the same encoder with different decoder components for various parsing task, including constituent parsing, dependency parsing and CCG parsing.



\bibliography{acl2017}
\bibliographystyle{acl_natbib}

\end{document}